\documentclass[letterpaper, 10 pt, journal, twoside]{IEEEtran}

\IEEEoverridecommandlockouts                              

\usepackage{amsmath} 
\usepackage{cases}
\usepackage{amssymb}  
\usepackage{hyperref}
\usepackage{multirow}
\usepackage{booktabs}
\usepackage{textcomp}
\usepackage{gensymb}
\usepackage{siunitx}
\usepackage{url}
\usepackage{mathtools}
\usepackage{subcaption}
\usepackage{bm}  
\usepackage{soul}
\usepackage{dblfloatfix}
\usepackage[noabbrev]{cleveref}
\usepackage[utf8]{inputenc}
\usepackage[english]{babel}

\usepackage{amsthm}
\usepackage{cases}
\usepackage{empheq}
\usepackage{tikz}

\newcommand\copyrighttext{%
  \footnotesize \textcopyright 2021 IEEE.  Personal use of this material is permitted.  Permission from IEEE must be obtained for all other uses, in any current or future media, including reprinting/republishing this material for advertising or promotional purposes, creating new collective works, for resale or redistribution to servers or lists, or reuse of any copyrighted component of this work in other works.}
\newcommand\copyrightnotice{%
\begin{tikzpicture}[remember picture,overlay]
\node[anchor=south,yshift=4pt] at (current page.south) {\parbox{\dimexpr\textwidth-\fboxsep-\fboxrule\relax}{\copyrighttext}};
\end{tikzpicture}%
}

\newtheorem{definition}{Definition}

\providecommand{\keywords}[1]{\textbf{\textit{Index terms---}} #1}

\usepackage{accents}

\crefrangeformat{equation}{eqs. #3(#1)#4--#5(#2)#6}

\newcommand{\vep}{\mbox{\boldmath $\epsilon$}}

\newcommand{\vth}{\mbox{\boldmath $\theta$}}

\newcommand{\vlambda}{\mbox{\boldmath $\lambda$}}

\newcommand{\vxi}{\mbox{\boldmath $\xi$}}
\newcommand{\vpi}{\mbox{\boldmath $\pi$}}

\newcommand{\vsigma}{\mbox{\boldmath $\sigma$}}
\newcommand{\vtau}{\mbox{\boldmath $\tau$}}

\newcommand{\vom}{\mbox{\boldmath $\omega$}}

\newcommand{\vGa}{\bm \Gamma}

\newcommand{\vLa}{\bm \Lambda}

\newcommand{\ve}{\bm e}
\newcommand{\vf}{\bm f}
\newcommand{\vg}{\bm g}
\newcommand{\vh}{\bm h}

\newcommand{\vp}{\bm p}
\newcommand{\vq}{\bm q}
\newcommand{\vr}{\bm r}

\newcommand{\vt}{\bm t}
\newcommand{\vu}{\bm u}
\newcommand{\vv}{\bm v}
\newcommand{\vw}{\bm w}
\newcommand{\vx}{\bm x}

\newcommand{\vB}{\bm B}
\newcommand{\vC}{\bm C}

\newcommand{\vI}{\bm I}

\newcommand{\vK}{\bm K}

\newcommand{\vM}{\bm M}

\newcommand{\vR}{\bm R}
\newcommand{\vS}{\bm S}
\newcommand{\vT}{\bm T}

\newcommand{\vY}{\bm Y}

\begin{document}

\title{Adaptive CLF-MPC With Application To Quadrupedal Robots}


\author{Maria Vittoria Minniti, Ruben Grandia, Farbod Farshidian, Marco Hutter 
\thanks{Manuscript received: September, 9, 2021; Accepted October, 30, 2021.}
\thanks{This paper was recommended for publication by Editor Clement Gosselin upon evaluation of the Associate Editor and Reviewers' comments.}
\thanks{This work was supported in part by the Swiss National Science Foundation through the National Centre of Competence in Digital Fabrication (NCCR dfab), in part by the Swiss National Science Foundation through the National Centre of Competence in Research Robotics (NCCR Robotics), and in part by the European Union’s Horizon 2020 research and innovation programme under grant agreement No 780883.} 
\thanks{All authors are with the Robotic Systems Lab, ETH Zurich, Zurich 8092, Switzerland {\tt\footnotesize \{mminniti\}@ethz.ch}}%
\thanks{Digital Object Identifier (DOI): \url{https://doi.org/10.1109/LRA.2021.3128697}}
}

\maketitle
\copyrightnotice{}

\markboth{IEEE Robotics and Automation Letters. Preprint Version. Accepted October, 2021}
{Minniti \MakeLowercase{\textit{et al.}}: Adaptive CLF-MPC with application to quadrupedal robots} 


\begin{abstract}
Modern robotic systems are endowed with superior mobility and mechanical skills that make them suited to be employed in real-world scenarios, where interactions with heavy objects and precise manipulation capabilities are required. For instance, legged robots with high payload capacity can be used in disaster scenarios to remove dangerous material or carry injured people. It is thus essential to develop planning algorithms that can enable complex robots to perform motion and manipulation tasks accurately. In addition, online adaptation mechanisms with respect to new, unknown environments are needed. In this work, we impose that the optimal state-input trajectories generated by Model Predictive Control (MPC) satisfy the Lyapunov function criterion derived in adaptive control for robotic systems. As a result, we combine the
stability guarantees provided by Control Lyapunov Functions
(CLFs) and the optimality offered by MPC in a unified adaptive framework, yielding an improved performance during the robot's interaction with unknown objects.
We validate the proposed approach in simulation and hardware tests on a quadrupedal robot carrying un-modeled payloads and pulling heavy boxes.
\end{abstract}


\section{INTRODUCTION}
\begin{figure}[htp]
   \centering
   \includegraphics[scale=0.3]{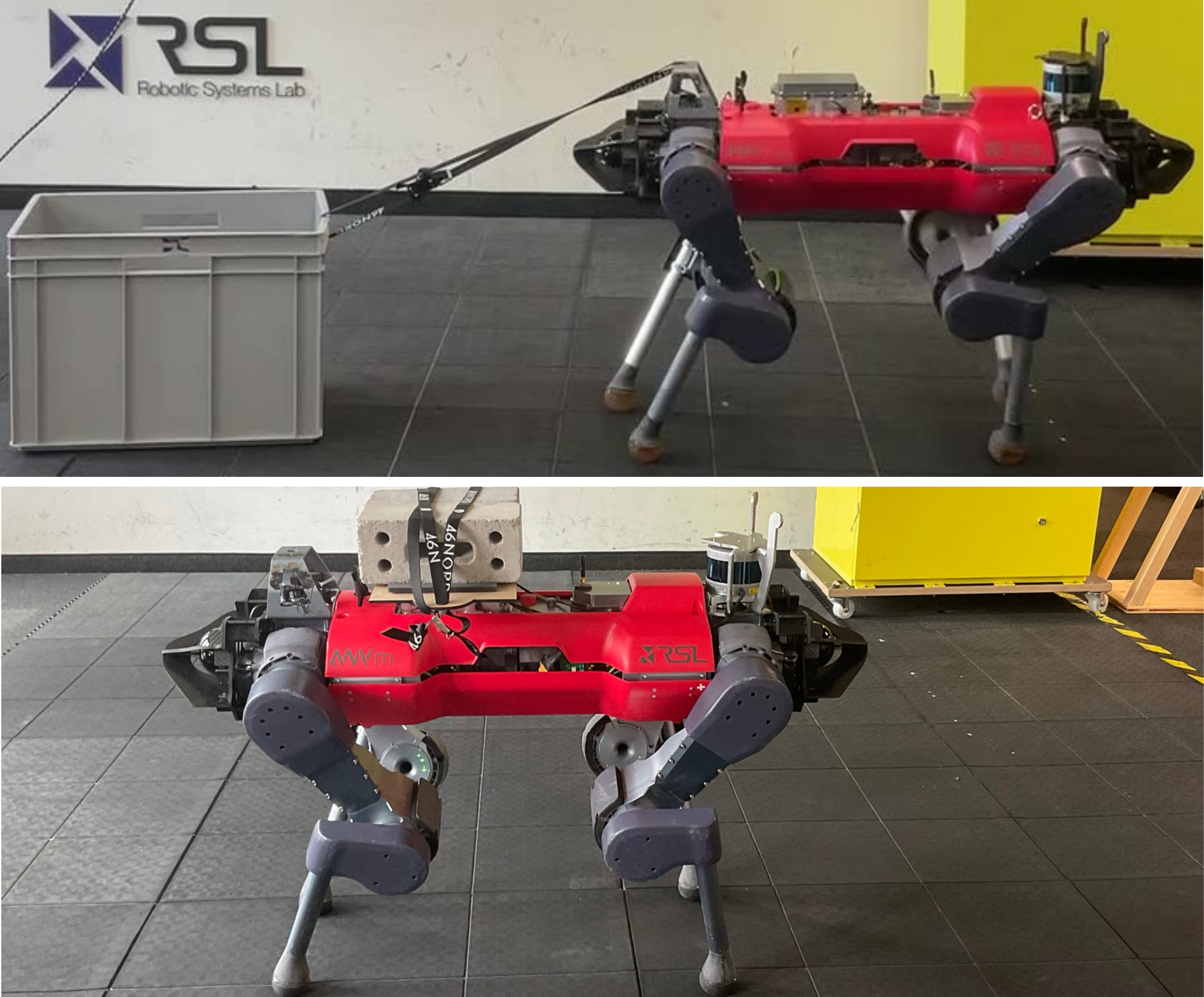}
   \caption{The quadrupedal robot ANYmal pulling a box of 21 kg (at the top), and holding a payload of 10.86 kg (at the bottom).} 
\label{fig:title_page}    
\end{figure}

\IEEEPARstart{O}ne major challenge for the real-world deployment of robots is the absence of a perfect model of all the objects that the robot should manipulate.
In addition, the model of the robot itself may be uncertain, since it is difficult to perfectly identify the dynamic parameters of all its components.
Based on this motivation, a fruitful, vast amount of literature has been dedicated to the development of adaptive control algorithms \cite{slotine1991applied} with a particular emphasis on direct approaches, such as the notable Slotine-Li adaptive controller for robot manipulators \cite{slotine1987adaptive}. 
However, when it comes to the control of poly-articulated mobile platforms, such as quadrupedal robots (Fig. \ref{fig:title_page}), it is necessary to devise methods that can reliably handle more complex, nonlinear performance objectives (such as foothold selection and end-effector tracking), as well as a set of safety constraints.
Thus, much analysis is currently being devoted to the study of approaches aimed at combining adaptive control laws with powerful learning and optimization algorithms \cite{boffi2021implicit}. In \cite{richards2021adaptive}, the disturbance experienced by the robot is parametrized through a neural network; the inner features parameters are meta-learned offline in closed-loop simulations of model ensembles under adaptive control. A deep network is used in \cite{joshi2020asynchronous} for quadrotor control; adaptation to different flight regimes is achieved using an MRAC law for the last layer weights. Research has also been conducted to add adaptive control stability guarantees in Quadratic Programs (QPs). Specifically for legged robots, an L1 adaptive controller is designed in \cite{chen2020adaptive}, \cite{nguyen20151} to track a nominal model based on a QP. For systems in strict feedback form, online adaptation and safety can be obtained by incorporating adaptive Control Lyapunov Functions (aCLFs) \cite{krstic1995control, moore2014adaptive} and adaptive Control Barrier Functions (aCBFs) in the QP formulation \cite{taylor2020adaptive}. Stochastic CBFs and stochastic CLFs have been used in QPs to obtain stability and safety conditions where model uncertainty is approximated via Bayesian model learning \cite{fan2020bayesian}.
CLFs have already been used as a powerful tool to synthesize direct adaptive controllers \cite{boffi2020learning}. In the context of infinite horizon optimal control, a direct adaptive controller can be obtained as the sum of the nominal optimal input, and an adaptive input that uses the optimal value function as a CLF \cite{lopez2021adaptive}.

Thanks to recent computational advances, Model Predictive Control (MPC) has shown real-time applicability on high-dimensional systems \cite{ding2019real}, \cite{farshidian2017real}.
However, the performance of nominal MPC degrades in the presence of model uncertainties. Robust MPC can cope with disturbances in a known set, although only few works have shown applications on real-world, robotic systems \cite{nubert2020safe}. Moreover, it has been demonstrated that online adaptation to the unknown parameters can provide MPC with robustness properties with respect to disturbances in a larger domain \cite{sinha2021adaptive}. In robotics, adaptive MPC controllers have mostly been designed in combination with system identification \cite{ding2019online, tournois2017online, haddadin2017robot}, or learning methods \cite{sun2021online}.
For mobile-manipulation problems, analysis on the combination of MPC with adaptive control has been conducted in \cite{minniti2021model}.

CLF stability criteria have also been added as constraints in MPC \cite{primbs2000receding}, \cite{yu2001comparison}, \cite{grandia2020nonlinear}, where stability guarantees are usually obtained with a properly designed terminal cost and terminal constraint \cite{rawlings2017model}. In the absence of such terminal conditions, which are complex to obtain for non-linear problems, MPC relies on the choice of the time horizon, with decreasing performance for smaller look-ahead. Incorporating CLF constraints within MPC allows to exploit the stability properties of CLF-QP controllers, and eases the tuning of the prediction horizon \cite{grandia2020nonlinear}. However, so far such advantages have not been shown in the presence of uncertainties in the robot non-linear MPC model.

Inspired by recent literature \cite{grandia2020nonlinear}, we propose here to combine CLFs with MPC, targeting robotic applications that involve interaction with unknown environments/objects. The contributions of this letter are the following:
\begin{itemize}
    \item We consider the Lyapunov condition derived in adaptive control for robotic manipulators \cite{slotine1987adaptive}, and we incorporate it as an inequality constraint to an MPC problem. Thus, the controller plans for an approximate system model, while ensuring asymptotic tracking in the presence of matched uncertainties. As a result, we combine the optimality of MPC with the advantages provided by CLFs. These include the stability guarantees and a reduced dependence on the prediction horizon. 
    \item We present the implementation of the proposed formulation on a floating-base model of a quadrupedal robot.
    \item We validate the proposed setup in a variety of simulations and real-world hardware experiments. The simulations aim to demonstrate the advantages of the proposed approach with respect to some baseline methods. These characterize themselves for the presence/absence of a terminal penalty, the awareness of the model mismatch, or the use of a different adaptation mechanism. The hardware tests showcase a quadrupedal robot carrying bricks and pulling heavy boxes.
\end{itemize}

\section{BACKGROUND}
\label{sec:background}
\subsection{Nonlinear Model Predictive Control}
\label{subsec:model_predictive_control}
We consider the following nonlinear optimal control problem (OCP):
\begin{subequations}
\begin{align}
     & \underset{\vu(\cdot)}{\text{minimize}} &&\int_0^T l(\vx(t), \vu(t))dt + \phi(\vx(T)) \label{eq:cost_function}\\
    &\text{subject to:} && \dot \vx = \vf (\vx) + \vB(\vx) \vu \label{eq:system_dynamics}\\
    & & &  \vx(0) = \vx_0 
    \label{eq:initial_cond} \\
    & & & \vg_1(\vx) = \mathbf{0}
    \label{eq:state_equalities} \\
    & & & \vg_2(\vx, \vu) = \mathbf{0}
    \label{eq:state_input_equalities} \\
    & & & \vh(\vx, \vu) \geq \mathbf{0} \label{eq:ineq_constraints} 
\end{align}
\end{subequations}
where $\vx(t) \in \mathcal{X}$, $\vu(t) \in \mathcal{U}$ are the state and input to the system, $l(\cdot)$ is an intermediate cost, $\phi(\cdot)$ is a terminal cost, and $T$ is the time horizon. The OCP is solved in real-time by updating the initial conditions \eqref{eq:initial_cond} with the measured state of the system. Eq. \eqref{eq:system_dynamics} describes the system dynamics, \Crefrange{eq:state_equalities}{eq:ineq_constraints} describe the state-only equality constraints, state-input equality constraints and inequality constraints, respectively.  Here, we consider an MPC implementation where inequality constraints \eqref{eq:ineq_constraints} are treated according to a relaxed-barrier method \cite{grandia2019feedback}: a relaxed log-barrier function is computed as a function of the inequality constraint, and added to the cost minimization.
Without loss of generality, we present the following derivation in the continuous-time domain. However, the proposed method does not depend on the specific implementation (e.g., continuous-time or discrete-time DDP, direct methods, etc...). In addition, in Sec. \ref{sec:ACLF-MPC_formulation} we assume that Eq. \eqref{eq:system_dynamics} represents a general mechanical system. In Sec. \ref{sec:implementation_for_a_quadrupedal_robot}, we apply this general formulation on the simplified kino-dynamic model of a quadrupedal robot.

\subsection{Control Lyapunov Functions}
\label{sec:control_lyapunov_functions}

We consider an error function $\vsigma : \mathcal{X} \times \mathbb{R}_+ \to \mathbb{R}^l$, with $l \leq \text{dim}(\mathcal{X})$. Let $\mathcal{C} = \{\vsigma(\vx, t), \vx \in \mathcal{X}, t \in \mathbb{R}_+\}\subseteq \mathbb{R}^l$. We assume that each component of $\vsigma$ has relative degree 1.
\begin{definition}[\textit{Class $\mathcal{K}$-functions} \cite{grandia2020nonlinear}]
A continuous function $\alpha : [0, a)\to \mathbb{R}_+$, with $a>0$, is a class-$\mathcal{K}$ function ($\alpha \in \mathcal{K}$) if $\alpha(0) = 0$ and $\alpha$ is strictly monotonically increasing. If $a = \infty$ and $\lim_{r\to\infty}\alpha(r) = \infty$, then $\alpha$ is said to be a class-$\mathcal{K}_{\infty}$ function.
\end{definition}

\begin{definition}[\textit{Control Lyapunov Funtions} \cite{taylor2019episodic, grandia2020nonlinear}] 
\label{def:CLF_definition}A continuously differentiable funtion $V:\mathbb{R}^l\to \mathbb{R}_+$ is said to be a Control Lyapunov Function (CLF) for \eqref{eq:system_dynamics} on $\mathcal{C}$ if there exist $\alpha_1, \alpha_2, \alpha_3 \in \mathcal{K}_{\infty}$ such that $\forall \vsigma \in \mathcal{C}$:
\begin{align}
    &\alpha_1(||\vsigma||) \leq V(\vsigma) \leq \alpha_2(||\vsigma||)  \\ 
    &\dot V(\vsigma, \vu) \leq - \alpha_3(||\vsigma||) \label{eq:lyapunov_derivative_constraint}
\end{align}
\end{definition}
The existence of a CLF $V(\vsigma)$ is a necessary and sufficient condition for the existence of a state-feedback controller $k: \mathcal{X} \to \mathcal{U}$ that makes $\vsigma$ globally asymptotically converge to $\mathbf{0}$. As shown in Sec. \ref{sec:ACLF-MPC_formulation}, for affine robotic systems and by an appropriate definition of $\vsigma$, convergence to the error surface $\vsigma=\mathbf{0}$ can also lead to the convergence of the full-system state $\vx$ to a desired state $\vx_d$. In QP-based controllers, a method to obtain stability guarantees is to add inequality \eqref{eq:lyapunov_derivative_constraint} to the set of constraints of the QP \cite{ames2013towards}. A natural extension for MPC problems consists in adding \eqref{eq:lyapunov_derivative_constraint} to the set of inequality constraints of Eq. \eqref{eq:ineq_constraints} \cite{primbs2000receding}, \cite{yu2001comparison}, \cite{grandia2020nonlinear}. 
We build from this idea for the following derivation.

\section{Adaptive clf-mpc for robotic systems}
\label{sec:ACLF-MPC_formulation}
\subsection{Adaptive Optimal Control Problem}
\label{sec:aclf-mpc_mechanical_systems}
We want to control mechanical systems in the form:
\begin{equation}
    \vM_n(\vq)\dot \vv + \vC_n(\vq, \vv)\vv + \vg_n(\vq) +
    \vtau_u(\vq, \vv, \dot \vv)= \vS(\vq) \vtau 
    \label{eq:robot_equations}
\end{equation}
where $\vq, \vv \in \mathbb{R}^n, \vtau \in \mathbb{R}^m$ are the generalized positions, velocities and torques of the robot, respectively, $\vS \in \mathbb{R}^{n\times m}$ is an actuator selection matrix, and $\vtau_u \in \mathbb{R}^n$ is an uncertainty term. The subscript $n$ refers to all the variables related to the \textit{nominal} model of the robot, which is usually acquired from offline identification. The subscript $u$ refers to the unknown terms, that may include external forces due to payloads or contact forces. We make the assumption that the uncertainty $\vtau_u$ is a \textit{matched} uncertainty, meaning that $\vtau_u\in \text{range}(\vS)$.
We assume that $\vtau_u(\vq, \vv, \dot \vv)$ can be parametrized linearly with respect to the constant, unknown parameters $\vpi_u \in \mathbb{R}^p$:
\begin{equation}
\label{eq:uncertainty_linearly_parametrizable}
    \vtau_u(\vq, \vv, \dot \vv) = \vY_u(\vq, \vv, \dot \vv)\vpi_u
\end{equation}
and that we can write:
\begin{equation}
    \vY_j(\vq, \vv, \dot \vv)\vpi_j = \vM_j(\vq)\dot \vv + \vC_j(\vq, \vv)\vv + \vg_j(\vq)
    \label{eq:uncertainty_linear_parametrization}
\end{equation}
where $j=n,u$, $\vM_j > 0$ and $\dot \vM_j - 2\vC_j$ is skew-symmetric. It can be verified that the assumptions on the unknown term $\vtau_u$ are not restrictive and that, for a quadrupedal robot, they cover the case where $\vtau_u$ includes gravitational and inertial effects due to un-modeled payloads at the main body.

Let $\tilde \vq, \tilde \vv$ be the generalized position and velocity errors with respect to some desired references $\vq_d(t), \vv_d(t)$. We define a \textit{composite error}  $\vsigma(\tilde \vq, \tilde \vv)$ as a function of the errors $\tilde \vq, \tilde \vv$ (and thus implicitly a function of $\vq, \vv, t$), with the property that $\tilde \vq \to \mathbf{0}$ when it evolves on the surface $\mathcal{S} = \{(\tilde \vq, \tilde \vv) \in \mathbb{R}^{2n} | \vsigma=\mathbf{0}\}$. In the literature \cite{slotine1987adaptive, culbertson2021decentralized}, $\mathcal{S}$ is usually called \textit{sliding surface} and referred to as $\vsigma=\mathbf{0}$. If $\vv = \dot \vq$, we can simply choose $\vsigma:=\tilde \vv + \vLa \tilde \vq$, with $\vLa > \mathbf{0}$. In addition, we define $\vv_r :=\vsigma + \vv$, $\vx := (\vq, \vv)$, $\vu := \vtau$. We propose the following adaptive CLF-MPC scheme (ACLF-MPC) for robotic systems:
\begin{subequations}
\begin{align}
     & \underset{\vu(\cdot)}{\text{minimize}} & &\int_0^T l(\vx(t), \vu(t))dt + \phi(\vx(T)) \label{eq:adapted_robot_cost_function} \\
    &\text{subject to:}  &&\dot \vx = \vf_a(\vx, \vu, \hat \vpi_u) \label{eq:adapted_system_dynamics}\\
    & & & \vx(0) = \vx_0 \notag \\
    & & &\vg_1(\vx) = \mathbf{0} 
    \label{eq:state_equalities_aclf_mpc_robot} 
    \\
    & & &\vg_2(\vx, \vu) = \mathbf{0}  
    \label{eq:state_input_equalities_aclf_mpc_robot} \\
    & & &\vh(\vx, \vu) \geq \mathbf{0} \label{eq:inequalities_aclf_mpc_robot}\\
    & & & h_{clf}(\vx, \vu, \hat \vpi_u) \geq 0 \label{eq:aclf_constraint_robot_compact}
\end{align}
\end{subequations}
where $\dot \vx = \vf_a$ in Eq. \eqref{eq:adapted_system_dynamics} is an adaptive system dynamics, parametrized by $\hat \vpi_u$.
We update the unknown parameters estimates $\hat \vpi_u$ according to:
\begin{equation}
    \label{eq:update_law_robot}
    \dot{\hat \vpi}_u = \vGa \vY_u^T(\vq, \vv, \vv_r, \dot \vv_r)\vsigma, \quad \vGa \in \mathbb{R}^{p\times p}, \vGa > 0,
\end{equation}
with a modified regressor $\vY_u$ obtained from the equation:
\begin{equation}
    \vY_u(\vq, \vv, \vv_r, \dot \vv_r)\vpi_u = \vM_u(\vq)\dot \vv_r + \vC_u(\vq, \vv)\vv_r + \vg_u(\vq).
\end{equation}
For the sake of brevity, in the following we omit the dependency of $\vY_u$ on $(\vq, \vv, \vv_r, \dot \vv_r)$.

\begin{figure}[t]
   \centering
   \includegraphics[scale=0.047]{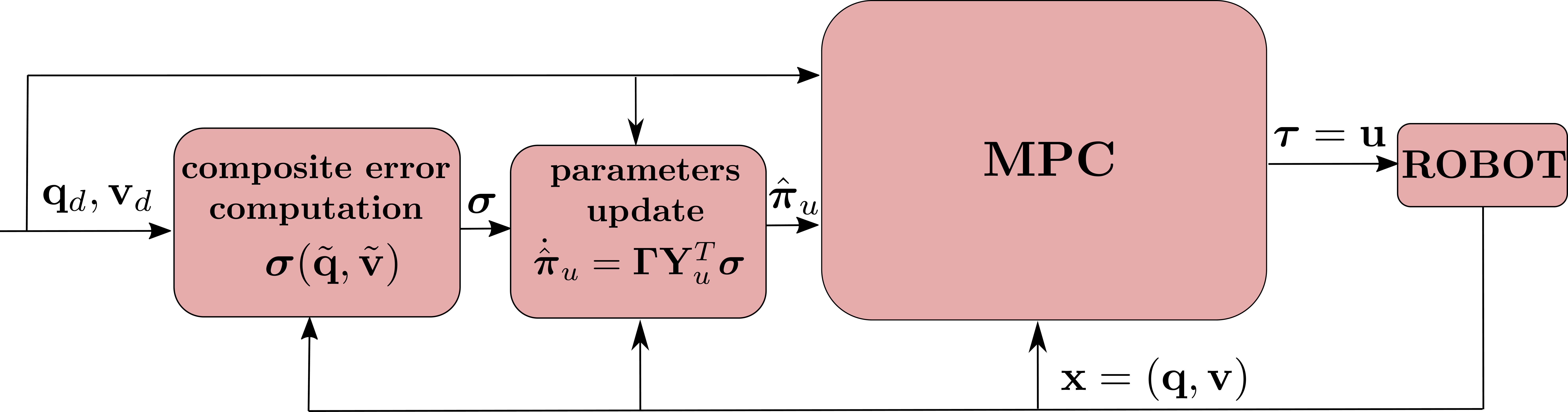}
   \caption{Block diagram of the main components of the ACLF-MPC method, as described in Sec. \ref{sec:aclf-mpc_mechanical_systems}.} 
\label{fig:ACLF-MPC_block_diagram}    
\end{figure}

\Crefrange{eq:state_equalities_aclf_mpc_robot}{eq:inequalities_aclf_mpc_robot} describe a set of safety constraints; for instance, in the case of a quadrupedal robot, they include friction cone constraints to avoid slippage and zero-velocity contraints at the contact points. In the adaptive dynamics of Eq. \eqref{eq:adapted_system_dynamics}, we define:
\begin{equation}
\scalebox{0.92}{$\dot \vv= \vM_n(\vq)^{-1}\left[-\vC_n(\vq, \vv)\vv -\vg_n(\vq) + \vS(\vq)\vtau - \vY_u\hat \vpi_u\right]$} \label{eq:adapted_robot_system_dynamics}
\end{equation}
Let $\vK_D \in \mathbb{R}^{n\times n}, \vK_D>\mathbf{0}$. The left-hand side of the constraint in Eq. \eqref{eq:aclf_constraint_robot_compact} has the form:
\begin{equation}
\label{eq:lhs_clf_constraint}
    h_{clf} = -\vsigma^T[- \vS(\vq)\vtau + \vY_n\vpi_n + \vY_u\hat \vpi_u]-\frac{1}{2}\vsigma^T\vK_D\vsigma 
\end{equation}
and will be derived explicitly in the following paragraph. A scheme of the main building blocks of the proposed method is provided in Fig. \ref{fig:ACLF-MPC_block_diagram}.

\subsection{CLF constraint and adaptive dynamics}
\label{sec:adaptive_clf_constraint_and_dynamics}
The inequality $h_{clf}\geq0$ in Eq. \eqref{eq:aclf_constraint_robot_compact} is set equal to the condition $\dot V \leq -W$, with the Lyapunov-like candidate $V$ and the function $W$ defined as:
\begin{align}
    &V(\vq, \vsigma, \tilde \vpi_u) := \frac{1}{2}\vsigma^T \vM(\vq)\vsigma + \frac{1}{2} \tilde \vpi_u^T \vGa^{-1}\tilde \vpi_u, \label{eq:adaptive_lyapunov_function} \\
    &W(\vsigma) := \frac{1}{2}\vsigma^T \vK_D \vsigma \notag
\end{align}
where $\tilde \vpi_u := \hat \vpi_u - \vpi_u$, and $\vM := \vM_n + \vM_u$.
We point out that $V$ is only positive definite in $\vsigma$ (and $\tilde \vpi_u$), and $\vsigma(\tilde \vq, \tilde \vv)$ is not a sufficient coordinate change for the full-dynamics of the system. However, the inequality $\dot V \leq - W$ still guarantees global asymptotic stability and boundedness of $\tilde \vq, \tilde \vv$. Indeed, if $\dot V \leq -W$, the error state $\tilde \vq, \tilde \vv$ converges to the sliding surface $\vsigma(\tilde \vq, \tilde \vv)=\mathbf{0}$. This yields $\tilde \vq \to \mathbf{0}$ as $t \to \infty$.

It can be verified that, if the estimates $\hat \vpi_u$ are updated according to Eq. \eqref{eq:update_law_robot}, $h_{clf}$ does not depend on the unknown parameters $\vpi_u$. To prove this, we note that the equations of motion \eqref{eq:robot_equations} can be written in an equivalent form:
\begin{align}
    &\vM(\vq)\dot \vsigma + \vC(\vq, \vv)\vsigma \label{eq:robot_equations_equivalent_form}\\ &=-\vS(\vq)\vtau +\vY_n(\vq, \vv, \vv_r, \dot \vv_r)\vpi_n   + \vY_u(\vq, \vv, \vv_r, \dot \vv_r)\vpi_u, \notag
\end{align}
with $\vC:=\vC_n+\vC_u$. Deriving $V$ along the trajectories of \eqref{eq:robot_equations_equivalent_form} yields:
\begin{align}
    \dot V &= \frac{1}{2}\vsigma^T \dot \vM(\vq) \vsigma + \vsigma^T \vM(\vq)\dot \vsigma + \tilde \vpi_u^T \vGa^{-1}\dot {\hat \vpi}_u \label{eq:lyapunov_derivative} \\
    &=\frac{1}{2}\vsigma^T[\dot \vM(\vq)-2\vC(\vq,\vv)]\vsigma + \vsigma^T[- \vS(\vq)\vtau \notag \\
    &\phantom{=}+ \vY_n\vpi_n + \vY_u\vpi_u] + \tilde \vpi_u^T \vGa^{-1}\dot {\hat \vpi}_u \notag \\
    &=\vsigma^T[- \vS(\vq)\vtau + \vY_n\vpi_n + \vY_u\hat \vpi_u] + \tilde \vpi_u^T[\vGa^{-1}\dot{\hat \vpi}_u\notag \\
    &\phantom{=}- \vY_u^T\vsigma] \notag\\
    &=\vsigma^T[- \vS(\vq)\vtau + \vY_n\vpi_n + \vY_u\hat \vpi_u] \label{eq:dotV}
\end{align} 
where the skew-symmetry of $\dot \vM-2\vC$ has been used, together with the update law \eqref{eq:update_law_robot}. Thus, re-formulating the inequality as $h_{clf} = - \dot V - W \geq 0$ yields Eq. \eqref{eq:lhs_clf_constraint}. 

Incorporating the constraint $h_{clf} \geq 0$ into the MPC problem forces the optimal input $\vtau$ to be a globally asymptotically stabilizing input for \eqref{eq:robot_equations}.
Theoretically, such a constraint is sufficient to guarantee global asymptotic stability. However, inequalities are usually treated as soft constraints \cite{grandia2019feedback}. On the other hand, the MPC dynamics \eqref{eq:adapted_system_dynamics} is a hard constraint. As a consequence, if \eqref{eq:adapted_system_dynamics} would only be based on the robot nominal model, it could conflict with the CLF constraint, leading to the latter not being satisfied. Thus, the term $\vY_u(\vq, \vv, \vv_r, \dot \vv_r) \hat \vpi_u$ is used in place of the true disturbance $\vY_u(\vq, \vv, \dot \vv) \vpi_u$ in the dynamic constraint of Eq. \eqref{eq:adapted_robot_system_dynamics}. The adaptive component $\vY_u \hat \vpi_u$ serves both as an estimate of the disturbance, and to generate the adaptation needed in the system dynamics to allow the satisfaction of the CLF constraint. Furthermore, this choice allows obtaining an MPC controller which is also a certainty equivalence controller \cite{sinha2021adaptive}. Indeed, Eq. \eqref{eq:adapted_robot_system_dynamics} is equivalent to incorporate the following two constraints in the MPC problem:
\begin{align}
    &\vS(\vq)\vw := \vS(\vq)\vtau - \vY_u(\vq, \vv, \vv_r, \dot \vv_r)\hat \vpi_u \label{eq:nominal_input}\\
    &\dot \vv = \vM_n(\vq)^{-1}[-\vC_n(\vq, \vv)\vv -\vg_n(\vq) +\vS(\vq)\vw] \label{eq:nominal_dynamical_system}
\end{align}
with the introduction of an auxiliary input $\vw \in \mathcal{U}$. Eq. \eqref{eq:nominal_dynamical_system} represents the nominal system, while Eq. \eqref{eq:nominal_input} can be solved exactly with respect to $\vtau$ because of the assumption of matched uncertainty. As a result, the optimal input $\vtau^*$ is given by the sum of a nominal input $\vw^*$, and an adaptive disturbance estimate.

\subsection{Convergence of the adaptive dynamics}
\label{sec:adaptive_dynamics_convergence}
In adaptive control, it is not required that the estimated parameters converge to the true values. However, for linear systems, there exist persistent excitation conditions that can be incorporated as MPC constraints to obtain such a convergence \cite{lu2019robust}. As pointed out in \cite{culbertson2021decentralized}, analogous simple conditions do not exist for nonlinear systems. Thus, here we aim to achieve approximate convergence of the adaptive uncertainty $\vY_u\hat \vpi_u$ to the true uncertainty $\vY_u \vpi_u$ by an appropriate cost function tuning. We refer again to the auxiliary input $\vw$ defined in Eq. \eqref{eq:nominal_input}. Because of the constraint $\eqref{eq:aclf_constraint_robot_compact}$, in closed-loop $\vsigma \to \mathbf{0}$, and thus the robot dynamics \eqref{eq:robot_equations_equivalent_form} reduces to:
\begin{equation*}
    \vS(\vq)\vtau \approx \vY_n \vpi_n + \vY_u\vpi_u \Rightarrow \vS(\vq)\vw + \vY_u\hat \vpi_u \approx \vY_n \vpi_n + \vY_u\vpi_u.
\end{equation*}
We suppose that the MPC cost is designed such that the difference $||\vS(\vq)\vw - \vY_n\vpi_n||$ is minimized. We are left with:
\begin{equation*}
\label{eq:adaptive_force_equality}
    \vY_u(\vq, \vv, \vv_r, \dot \vv_r)\hat \vpi_u \approx \vY_u(\vq, \vv, \vv_r, \dot \vv_r) \vpi_u + \vep
\end{equation*}
where $\vep$ is a residual error from the minimization of $||\vS(\vq)\vw - \vY_n\vpi_n||$. If such an error is small, the adaptive term $\vY_u \hat \vpi_u$ converges to the true uncertainty $\vY_u \vpi_u$. In the absence of persistently exciting references, this does not imply $\hat \vpi_u \approx \vpi_u$. However, it implies that the MPC dynamics approximately converges to the true dynamics.

\section{Implementation for a quadrupedal robot}
\label{sec:implementation_for_a_quadrupedal_robot}
\begin{figure}[t]
   \centering
   \includegraphics[scale=0.50]{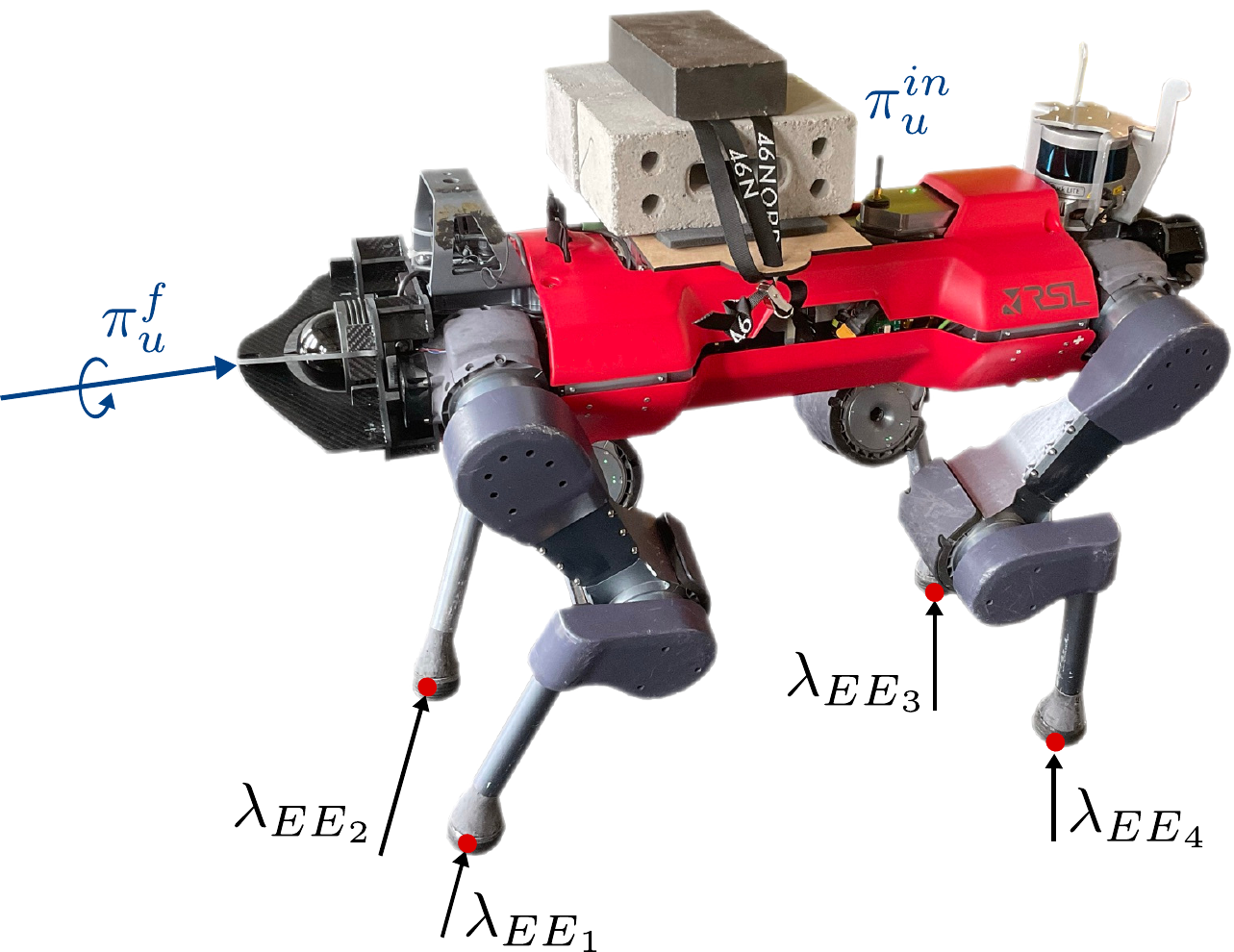}
   \caption{Illustration of the considered scenario for a quadrupedal robot. The MPC control inputs for the single rigid-body model are the feet contact forces $\vlambda_{{EE}_i}$. The uncertain parameters are due to un-modeled payloads ($\vpi_u^{in}$) and constant wrenches acting on the base ($\vpi_u^f$).} 
\label{fig:stilistic_illustration}    
\end{figure}

In this section, we describe the implementation of the MPC problem given in \Crefrange{eq:adapted_robot_cost_function}{eq:update_law_robot} for the 6-dimensional floating-base model of a quadrupedal robot interacting with un-modeled objects at its base (Fig. \ref{fig:stilistic_illustration}).

\subsection{Adaptive dynamics and constraints}
\label{subsec:adaptive_modification}
To derive the adaptive modification of the MPC dynamics, we base ourselves on the kino-dynamic model described in \cite{grandia2019feedback}. The adaptive equations of motion are given by:
\begin{numcases}{}
\dot \vp = \vv_p \label{eq:com_position_dynamics}\\
\dot \vth = \vT(\vth)\vom \label{eq:base_rotation_dynamics} \\
\dot \vv_p = \vg + \frac{1}{m}(\sum\nolimits_{i=1}^4\vR_{WB}\vlambda_{EE_i} - \vR_{WB} \vf_u) \label{eq:quadruped_adaptive_linear_dynamics}\\
\dot \vom = \vI^{-1}(-\vom\times \vI \vom + \sum\nolimits_{i=1}^4\vr_{EE_i}\times \vlambda_{EE_i} - \vt_u) \label{eq:quadruped_adaptive_rotational_dynamics} \\
\dot \vth_j = \vxi_j \label{eq:quadruped_joints_kinematics}
\end{numcases}
where $\vp, \vth \in \mathbb{R}^3$ are the position of the robot center of mass (CoM) in world frame $\{W\}$ and the base Euler angles, respectively. $\vT(\vth)$ is the mapping from the base angular velocity $\vom$, expressed in base frame $\{B\}$, and the Euler angles time derivative $\dot \vth$. $\vI$ and $m$ are the moment of inertia about the CoM and the mass of the robot, respectively. $\vr_{EE_i}$ is the position of the foot $i$ with respect to the CoM, and $\vth_j$ are the legs' joint positions. The MPC control input is given by $\vu = (\vxi_j, \vlambda_{{EE}_i})$, where $\vxi_j$ are the legs' joint velocities and $\vlambda_{{EE}_i}$ is the reference contact force for the $ith$-foot, expressed in base frame. The kino-dynamic model described in Eqs. \eqref{eq:com_position_dynamics}-\eqref{eq:quadruped_joints_kinematics} is composed of a free-floating, single rigid-body model and a kinematic model for the legs' joints. Here, we concentrate on the problem of tracking a desired pose (center of mass position and base orientation) for the single rigid-body subsystem (Eqs. \eqref{eq:com_position_dynamics}-\eqref{eq:quadruped_adaptive_rotational_dynamics}) which has the properties described in Sec. \ref{sec:aclf-mpc_mechanical_systems}. Let $\vq := (\vp, \vth) \in \mathbb{R}^6$ and $\vv := (\vv_p, \vom) \in \mathbb{R}^6$, respectively. 
We define an \textit{adaptive wrench} (combined force and torque) acting on the robot base:
\begin{equation}
     \begin{bmatrix} \vf_u \\ \vt_u 
    \end{bmatrix}= \vY_u(\vq, \vv, \dot \vv) \vpi_u = \vY_u^{in}(\vq, \vv, \dot \vv) \vpi_u^{in} + \vY_u^{f}(\vq, \vv) \vpi_u^{f} \label{eq:adaptive_wrench_quadruped}
\end{equation}
In Eq. \eqref{eq:adaptive_wrench_quadruped}, $\vpi_u^{f} \in \mathbb{R}^{6}$ includes constant forces and torques acting on the robot, while $\vpi_u^{in} \in \mathbb{R}^{10}$ is a vector of unknown inertial parameters (a combination of the mass of the unknown payload, center of mass and inertia).
 $\vY_u^{in}$ can be derived through the standard procedure for the Slotine-Li regressor \cite{siciliano2010robotics}, and distinguishing between the robot nominal parameters $\vpi_n^{in}$ and those of the payload $\vpi_u^{in}$.
 Since a legged robot is under-actuated in many walking scenarios \cite{grandia2019feedback}, the minimization described in Sec. \ref{sec:adaptive_dynamics_convergence} is not perfectly achievable for the full 6-dimensional set of generalized coordinates. Thus, here we impose it only for the translational part of the floating-base model. This is equivalent to designing the cost function so that it equally distributes the robot weight among the feet in contact, such that: $||\sum_{i=1}^4 \vR_{WB} \vlambda_{{EE}_i} - m \vv_p^r +m\vg - \vR_{WB} \vf_u||\approx 0$.
 
 \Crefrange{eq:state_equalities_aclf_mpc_robot}{eq:inequalities_aclf_mpc_robot} include end-effector velocity constraints for swing and stance legs, as well as friction cone constraints for the feet in contact; we refer the reader to \cite{grandia2019feedback} for a more detailed description of the nominal model and constraints.

As described in Sec. \ref{sec:aclf-mpc_mechanical_systems}, we make the assumption that the parametric uncertainty $\vY_u \vpi_u$ is in a matched form. In practice, for a quadrupedal robot this requires that the robot has at least three feet in contact, or that the disturbance $\vY_u \vpi_u$ acts along a controllable direction (for instance, during a trotting gait, only the angle around the line of the two contact points is not controllable).

\subsection{Sliding surface for pose control}
\label{subsec:sliding_pose_control}

The implementation of the adaptive Lyapunov constraint requires the definition of the composite error $\vsigma := (\vsigma_l, \vsigma_o)$ for the floating-base linear and rotational dynamics. For the linear part, we can simply use $\vsigma_l := \tilde \vv_p + \vLa_l \tilde \vp$. Indeed, the position error $\tilde \vp$ evolves on the surface $\vsigma_l = \mathbf{0}$ with an asymptotically stable dynamics. As discussed in \cite{culbertson2021decentralized}, the implementation of an adaptive controller for the rotational dynamics requires the choice of a suitable rotation error. Here, we use the quaternion error as defined in \cite{siciliano2010robotics}:
\begin{equation}
    \ve_o := \eta \vep_d - \eta_d \vep - {\vep_d}^\times\vep 
\end{equation}
where $\mathcal{Q} = (\eta, \vep)$, $\mathcal{Q}_d = (\eta_d, \vep_d)$ are the base actual and desired quaternions, respectively, and $(\cdot)^\times$ is the skew operator. As demonstrated in \cite{siciliano2010robotics}, if we define $\vsigma_o := \tilde \vom + \vLa_o \ve_o$, we have that $\ve_o \to \mathbf{0}$ on the surface $\vsigma_o = \mathbf{0}$. 

\subsection{Conversion to torque commands}
\label{subsec:conversion_to_torque_commands}
For a robotic arm, the MPC control torques $\vtau$ can be directly commanded to the robot (Fig. \ref{fig:ACLF-MPC_block_diagram}). However, for a floating-base system, a step is needed to convert MPC optimal control inputs to torque commands. Here, desired contact forces, along with desired accelerations obtained from the forward simulation of Eqs. \eqref{eq:quadruped_adaptive_linear_dynamics}, \eqref{eq:quadruped_adaptive_rotational_dynamics} are tracked by a hierarchical whole-body QP controller \cite{bellicoso2016perception}, where the learned dynamics $\vY_u\hat \vpi_u$ is also compensated to take care of the model mismatch.

\section{Results}
\label{sec:experiments}
In this section, we validate the proposed approach in simulation and hardware tests on the quadrupedal robot ANYmal (Fig. \ref{fig:title_page}). A video showcasing the results accompanies this letter \footnote{Available at \url{https://youtu.be/Gu2mfAAvT0A}.}. For the underlying MPC computations, we use a Multiple Shooting algorithm provided by the OCS2 toolbox \cite{OCS2}. The problem is formulated as an OCP for switched systems \cite{farshidian2017real}, where the CLF constraint in Eq. \eqref{eq:aclf_constraint_robot_compact} depends on different contact conditions and is thus affected by the transitions between subsystems. During the hardware tests, all computations run on a single on-board PC (Intel i7-8850H, 2.6 GHz, hexa-core 64-bit) with the MPC solver running at 100 Hz with a time horizon of 1 s and the whole-body QP controller and state estimation running at 400 Hz. In all the proposed tests, we assume that the regressor related to the unknown parameters is as in \eqref{eq:adaptive_wrench_quadruped}. Unless otherwise stated, the uncertainties include both un-modeled inertial parameters $\vpi_u^{in} \in \mathbb{R}^{10}$, and constant disturbance forces applied on the base $\vpi_u^f \in \mathbb{R}^3$. These are either due to forces generated on purpose in the simulation environment (Sec. \ref{subsec:momentum_observer_comparison}), to the static friction torque present in the actuators (Sec. \ref{subsec:brick_carrying_experiment}), or to the static/dynamic force between the ground and the object that the robot carries (Sec. \ref{subsec:cart_experiment}).

\subsection{Comparative analysis}
\label{subsec:comparative_analysis}
We perform a comparative analysis of different methods which are feasible to be applied on the high-dimensional model of a quadrupedal robot. The analysis is conducted in a physics simulation, where we can precisely quantify the amount of model mismatch. We assume that the MPC model under-estimates the actual ANYmal mass of 20 kg ($\approx$40$\%$ of the nominal mass), and that the true center of mass is displaced by 0.3 m along the base x-direction from the center of mass of the MPC model.
We test the following five baselines:
\begin{enumerate}
    \item ACLF-MPC
    \item ACLF-MPC without terminal cost
     \item CLF-MPC
    \item Perfect-model MPC with terminal cost
    \item Perfect-model MPC without terminal cost
\end{enumerate}
The controllers 1) and 2) correspond to the presented method, tested with and without a terminal cost. This is set equal to the value function of an LQR, obtained from the linearization of the original problem. Indeed, although the terminal cost has been presented as part of the proposed method in Sec. \ref{sec:ACLF-MPC_formulation}, here we show that the CLF constraint helps reducing the controller dependency on such terminal component. The CLF-MPC without adaptive ability is tested in 3), where the stability constraint is obtained from the CLF of the nominal model.

In terms of tracking performance, ideal results are attainable with an MPC problem that is perfectly aware of the model mismatch. Baselines 4) and 5) correspond to this ideal case, which does not include a CLF constraint and where the model mismatch is known to the controller.

All the methods employ the same intermediate cost function, which was tuned so that the best performance is obtained in the nominal case. For the ACLF-MPC, the adaptation gain matrix is chosen as $\vGa = \text{diag}(\vGa_m, \vGa_{com}, \vGa_I)$, where $\vGa_m = 5.0$,  $\vGa_{com} = I_{3 \times 3}$, and $\vGa_{I} = 0.01 I_{6 \times 6}$ relate to the uncertainties in the robot mass, combined center of mass and rotational inertia, respectively. The initial values of the estimated parameters are set to zero. The tuning gains $\vLa_l, \vLa_o$ were chosen as $\vLa_l$ = $\vLa_o$ = $5 I_{3\times3}$, while ${\vK_D} = \text{diag}(50,50,50,80,80,80)$. During the simulation, we command the robot to stand while following a desired center of mass position and base orientation trajectory. The reference trajectory consists of a sequence of step references for the center of mass vertical component, and the base pitch and roll angles. 
To validate the advantages that the ACLF-MPC also provides with respect to the prediction horizon tuning, we repeat the same test under a time horizon of 1 s and 0.5 s, respectively. 
The root mean square linear and rotational errors (RMSE) are reported in Tab. \ref{table:comparative_analysis_time_horizon_1.0_sec} and \ref{table:comparative_analysis_time_horizon_0.5_sec}.

\begin{figure}[t]
   \centering
   \includegraphics[scale=0.18]{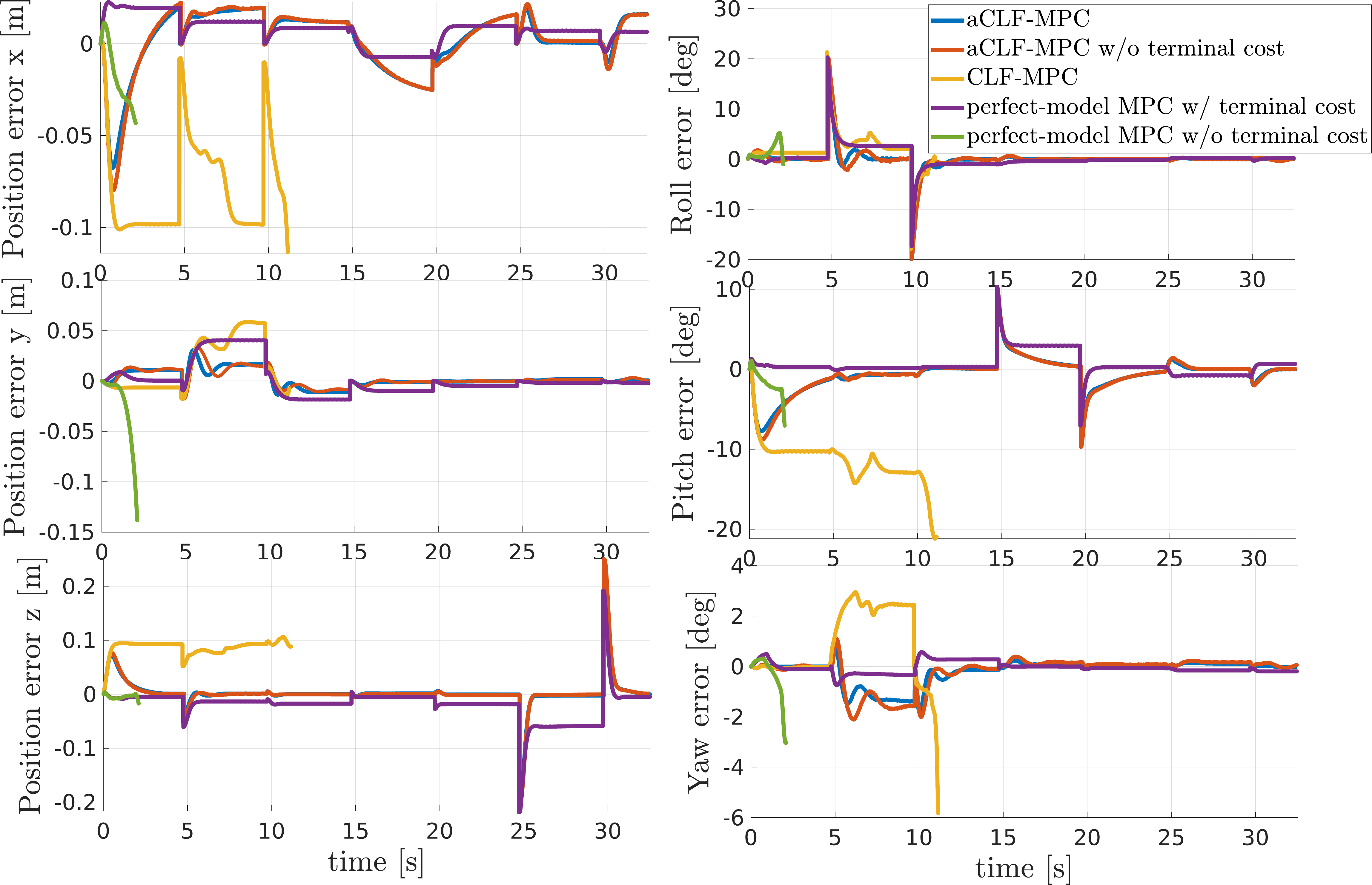}
   \caption{Position and rotation errors for a simulation test with a quadrupedal robot under the five baseline MPC controllers introduced in Sec. \ref{subsec:comparative_analysis}, with a time horizon of 0.5 s.} 
\label{fig:comparative_analysis_time_horizon_0.5_sec}    
\end{figure}

\begin{table}[t]
\centering
\caption{Linear and angular RMSE errors for a simulation test with a quadrupedal robot under the five baseline methods described in Sec. \ref{subsec:comparative_analysis}, with a time horizon of 1 s.}
\begin{tabular}{lrrrrr}
\hline
Method          & \multicolumn{1}{c}{1} & \multicolumn{1}{c}{2} & \multicolumn{1}{c}{3} & \multicolumn{1}{c}{4} & \multicolumn{1}{c}{5}\\ \hline
Linear RMSE [m]      & \textbf{0.036}                    & \textbf{0.036}                           & 0.10                    & 0.04 & 0.04                        \\
Angular RMSE [deg]                 & 3.23                     & 3.32                   & 10.2                 & \textbf{3.12}  & 3.14                 \\ \hline
\end{tabular}
\label{table:comparative_analysis_time_horizon_1.0_sec}
\end{table}

\begin{table}[t]
\centering
\caption{RMSE for a simulation test with a quadrupedal robot under the five baseline MPC controllers described in Sec. \ref{subsec:comparative_analysis}, with a time horizon of 0.5 s. A bar - indicates that the controller failed.}
\begin{tabular}{lrrrrr}
\hline
Method          & \multicolumn{1}{c}{1} & \multicolumn{1}{c}{2} & \multicolumn{1}{c}{3} & \multicolumn{1}{c}{4} & \multicolumn{1}{c}{5}\\ \hline
Linear RMSE [m]      & \textbf{0.035}                     & 0.037                           & -                    & 0.04 & -                         \\
Angular RMSE [deg]                 & 3.16                     & 3.56                   & -                   & \textbf{2.62}   & -                 \\ \hline
\end{tabular}
\label{table:comparative_analysis_time_horizon_0.5_sec}
\end{table}
With a time horizon of 1.0 s, the ACLF-MPC formulation performs comparatively well as the perfect-model MPC controller. In particular, the position tracking error even slightly improves with the proposed formulation with respect to the theoretical perfect-model baseline. In contrast, the ACLF-MPC outperforms the CLF-MPC that does not use any adaptation mechanism.

Interesting conclusions can be drawn from the behaviour of the five MPC controllers under a smaller time horizon (Fig. \ref{fig:comparative_analysis_time_horizon_0.5_sec} and Table \ref{table:comparative_analysis_time_horizon_0.5_sec}). In fact, the system becomes unstable under the perfect-model MPC controller that does not use a terminal cost. In addition, the CLF-MPC problem without adaptation becomes unfeasible due to the planned contact forces not adapting to the unknown payload distribution. On the contrary, the ACLF-MPC controller determines a closed-loop stable behavior even without the need for a terminal cost.

\subsection{Comparison with an online adaptation method}
\label{subsec:momentum_observer_comparison}
As an additional baseline comparison, we consider an adaptive method that estimates external forces and torques exerted on the robot base according to a generalized momentum approach, based on \cite{haddadin2017robot}. As in the proposed method, the estimated wrench is compensated online in the MPC dynamics. Such an approach is purely based on system identification, while our method is driven by the convergence of the pose tracking error. We consider a Gazebo simulation scenario where the robot walks up a slope with a statically stable gait, while it is subject to a constant disturbance force $F$ acting in the opposite direction to the robot motion. Here, walking on an inclined surface is proposed as a representative scenario where the base orientation tracking has a relevant influence on the balance of the robot. We gradually increase the value of the constant external force to validate which of the two methods can tolerate the largest amount of un-modeled disturbance. 
We find that, under the momentum observer method, the robot loses its balance along the slope when $F \geq 70N$, while the robot remains stable under the proposed approach. In fact, the momentum observer approach does not include the CLF constraint in the MPC problem. Consequently, the contact forces are computed based only on the cost function minimization, which tends to distribute the forces equally among the feet in contact. In a scenario where an external force is applied on the robot in the direction opposite to the motion, a torque around the base is generated, leading to a larger orientation error. On the other hand, the ACLF-MPC method explicitly adds an additional tracking requirement, which is embedded in the Lyapunov constraint. This in turn allows the robot to maintain balance while walking on the inclined surface.

\subsection{Base tracking under heavy bricks}
\label{subsec:brick_carrying_experiment}
\begin{figure}[t]
   \centering
   \includegraphics[scale=0.33]{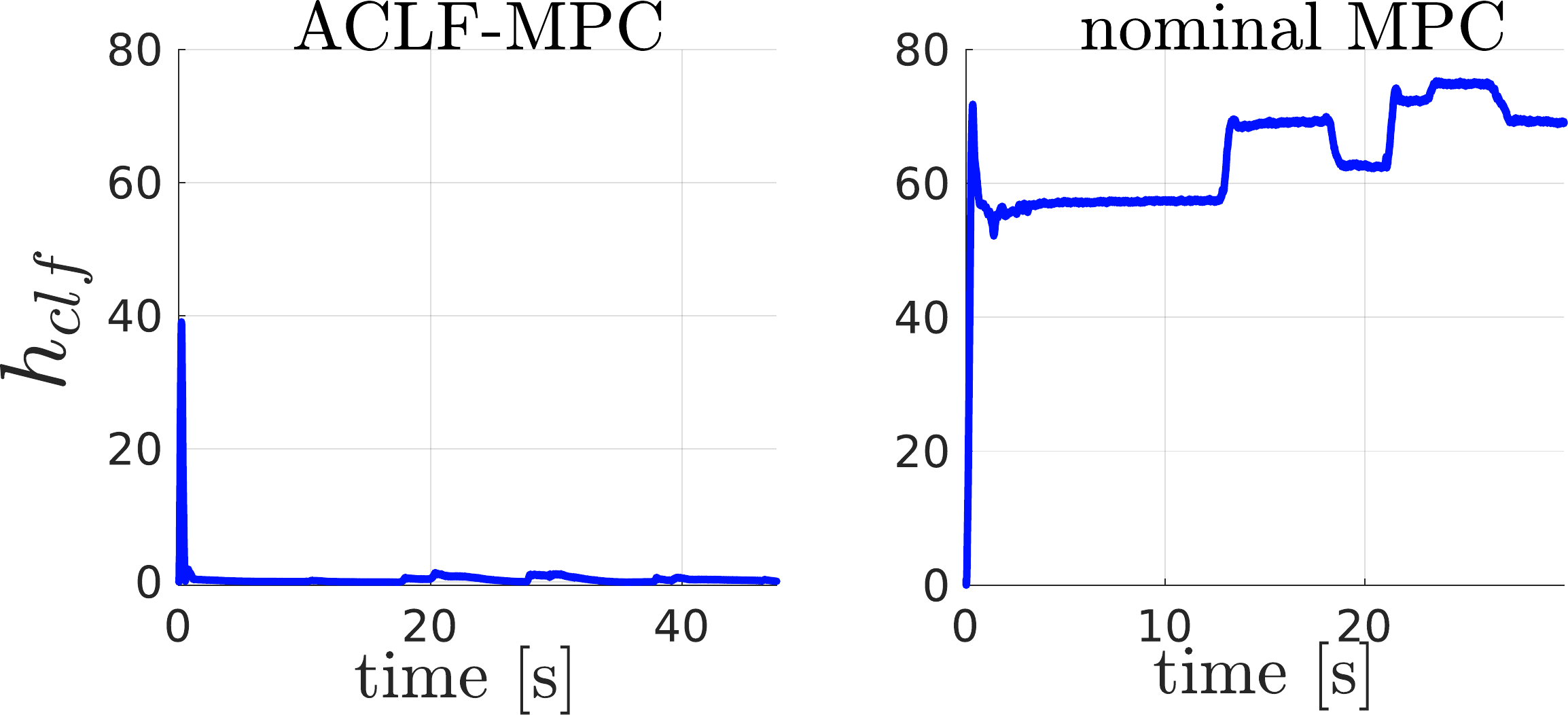}
   \caption{CLF constraint for the experiment described in Sec. \ref{subsec:brick_carrying_experiment}. The plots refer to two different tests where the robot is controlled with the ACLF-MPC approach (on the left), and with the nominal MPC (on the right).}
\label{fig:adaptive_clf_constraint_varying_com}    
\end{figure}

\begin{figure}[t]
   \centering
   \includegraphics[scale=0.27]{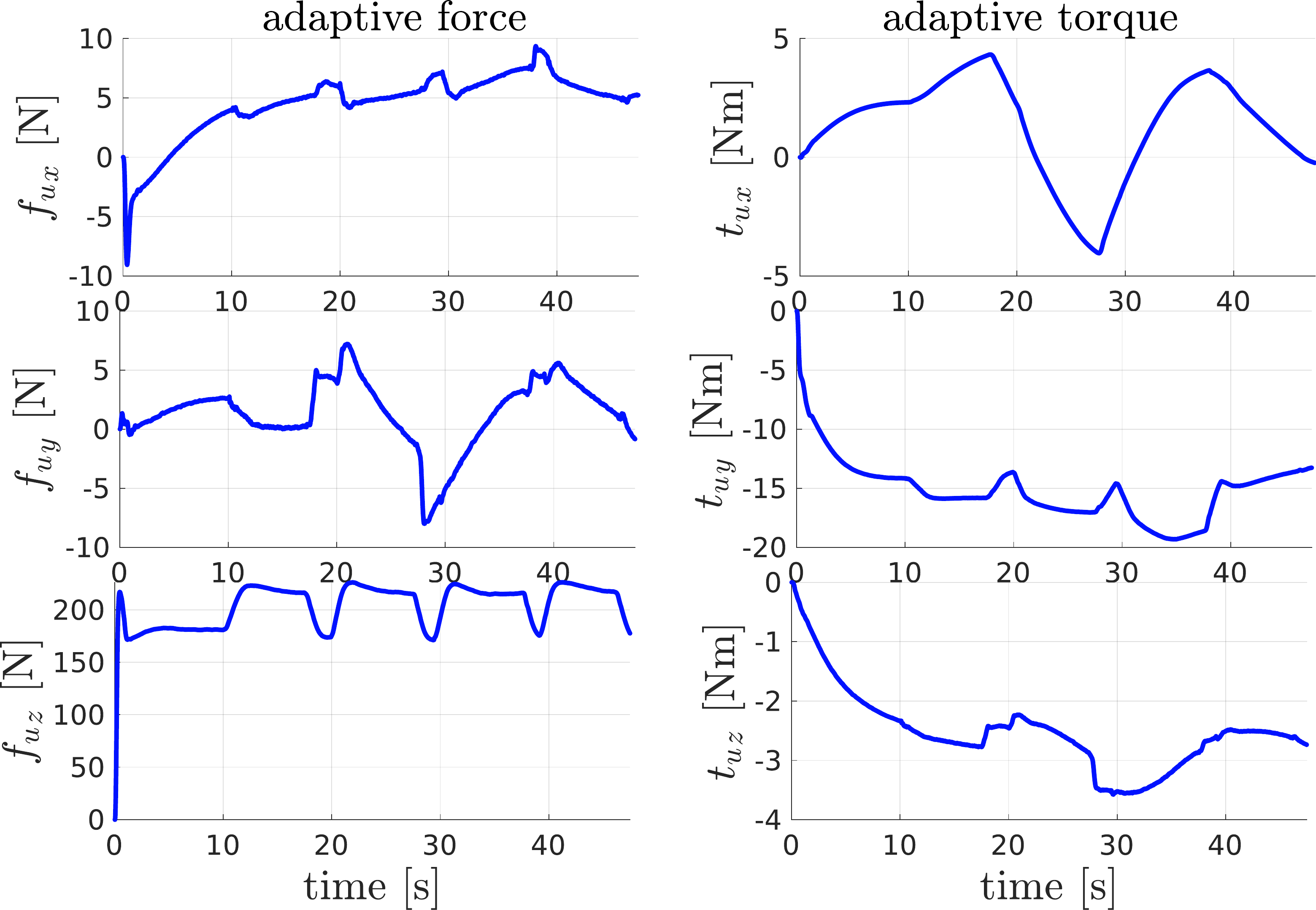}
   \caption{Adaptive force (on the left) and torque (on the right) for an experiment where the robot holds two un-modelled bricks, while the system center of mass is modified by placing a third brick at different positions on the base.}
\label{fig:adaptive_wrench_varying_com}    
\end{figure}

We also validate how the ACLF-MPC formulation performs in a set of real-world hardware experiments. Here, we place two heavy bricks on top of the robot; the weight of each brick is 5.43 kg. Moreover, we add a third brick of 3.4 kg at random positions on top of the base, in such a way as to cause a perturbation to the center of mass of the system. An additional source of modeling errors comes from the effects that the payload has on the dynamics of legs' joints, and that is neglected in the MPC kino-dynamic model. Initially, we command the robot to stand on four feet while keeping a desired constant base position and orientation. We repeat this experiment under the baseline nominal MPC controller (i.e., based on the robot nominal model), and the ACLF-MPC approach. As a compact measure of the tracking performance, in Fig. \ref{fig:adaptive_clf_constraint_varying_com} we show the constraint $h_{clf}$ from Eq. \eqref{eq:aclf_constraint_robot_compact}. Under the nominal MPC controller, $h_{clf}$ settles at a non-zero value. On the contrary, under the proposed controller, the value of the constraint converges to 0. This implies that the tracking error converges to the surface $\vsigma = \mathbf{0}$. As described in Sec. \ref{sec:ACLF-MPC_formulation}, convergence of the tracking error is achieved by the combination of a soft inequality constraint, and by compensating the adaptive wrench $\vY_u \hat \vpi_u = (\vf_u, \vt_u )$ in the MPC dynamics (Eq. \eqref{eq:adapted_robot_system_dynamics}). The latter is displayed in the plots of Fig. \ref{fig:adaptive_wrench_varying_com}. These show how all the components of the adaptive force and torque contribute to compensate for the center of mass perturbation.

As shown in the attached video, we also send position and orientation references to the robot during walking. We repeat such experiment with two bricks on top of the robot, under the nominal MPC controller and the ACLF-MPC controller, respectively. Under the nominal MPC controller, the robot exhibits an unbalanced behavior, due to not being able to track the desired pose commands (especially with respect to the z position and pitch angle). On the other hand, it can be seen from the video that, under the proposed formulation, the robot walks stably.

\subsection{Box pulling}
\label{subsec:cart_experiment}
\begin{figure}[t]
   \centering
   \includegraphics[scale=0.21]{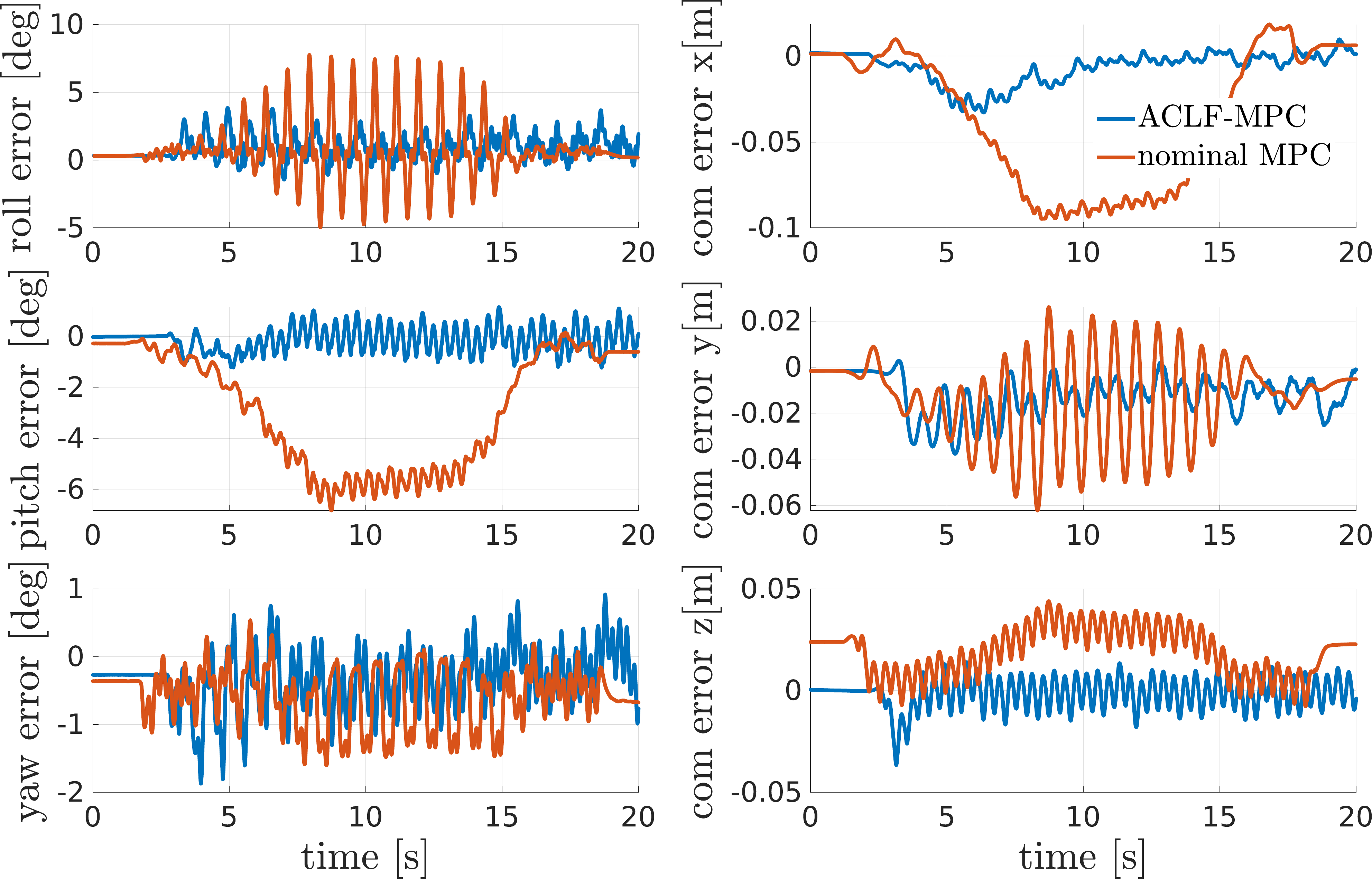}
   \caption{Base rotation and center of mass position errors for an experiment where the robot pulls an un-modelled box of 16.17 kg.}
\label{fig:cart_pulling_2_bricks_tracking_errors}    
\end{figure}

\begin{figure}[t]
   \centering
   \includegraphics[scale=0.21]{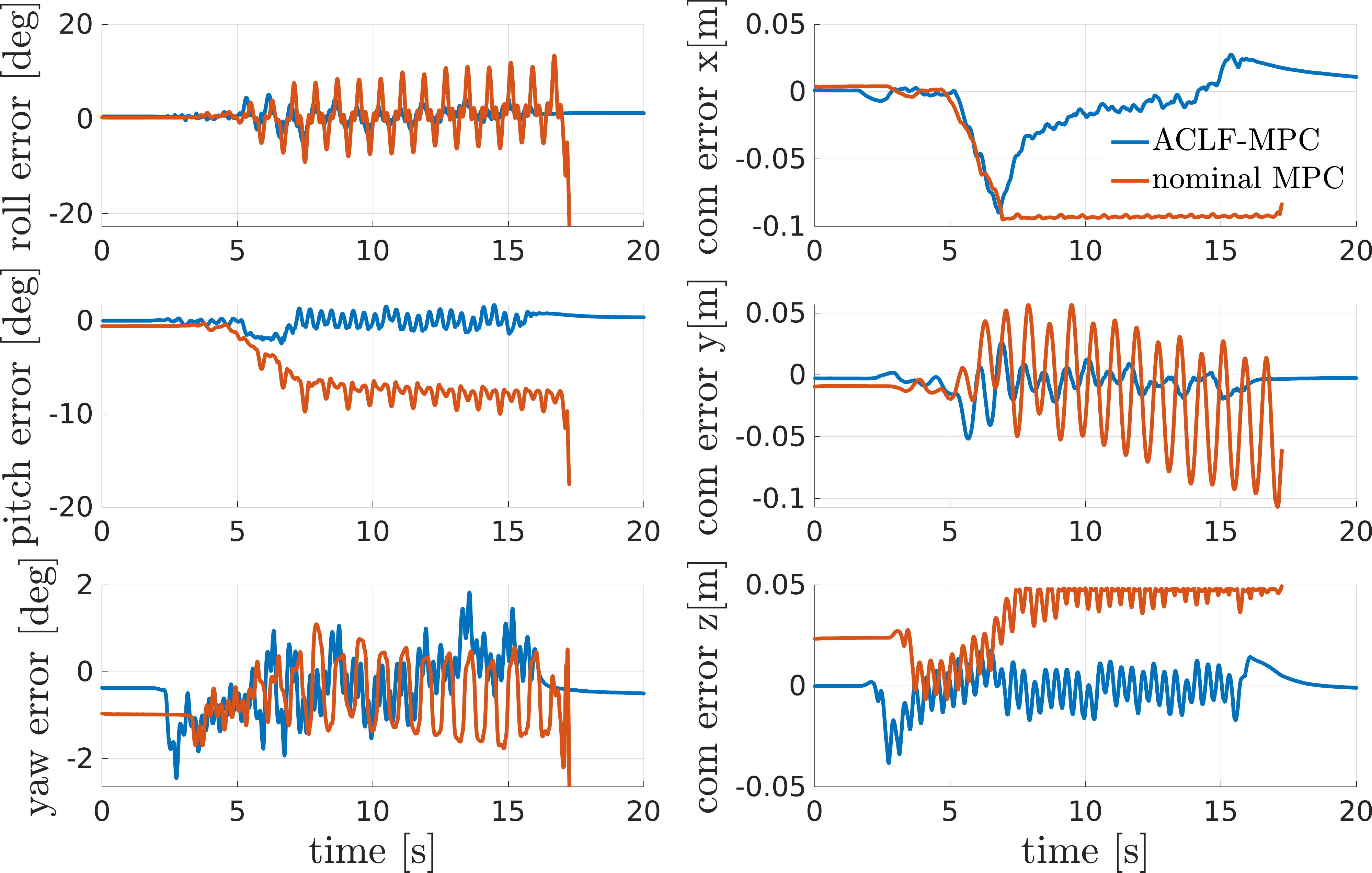}
   \caption{Base rotation and center of mass position errors for an experiment where the robot pulls an un-modelled box of 21.03 kg.}
\label{fig:cart_pulling_3_bricks_tracking_errors}    
\end{figure}

\begin{figure}[t]
   \centering
   \includegraphics[scale=0.2]{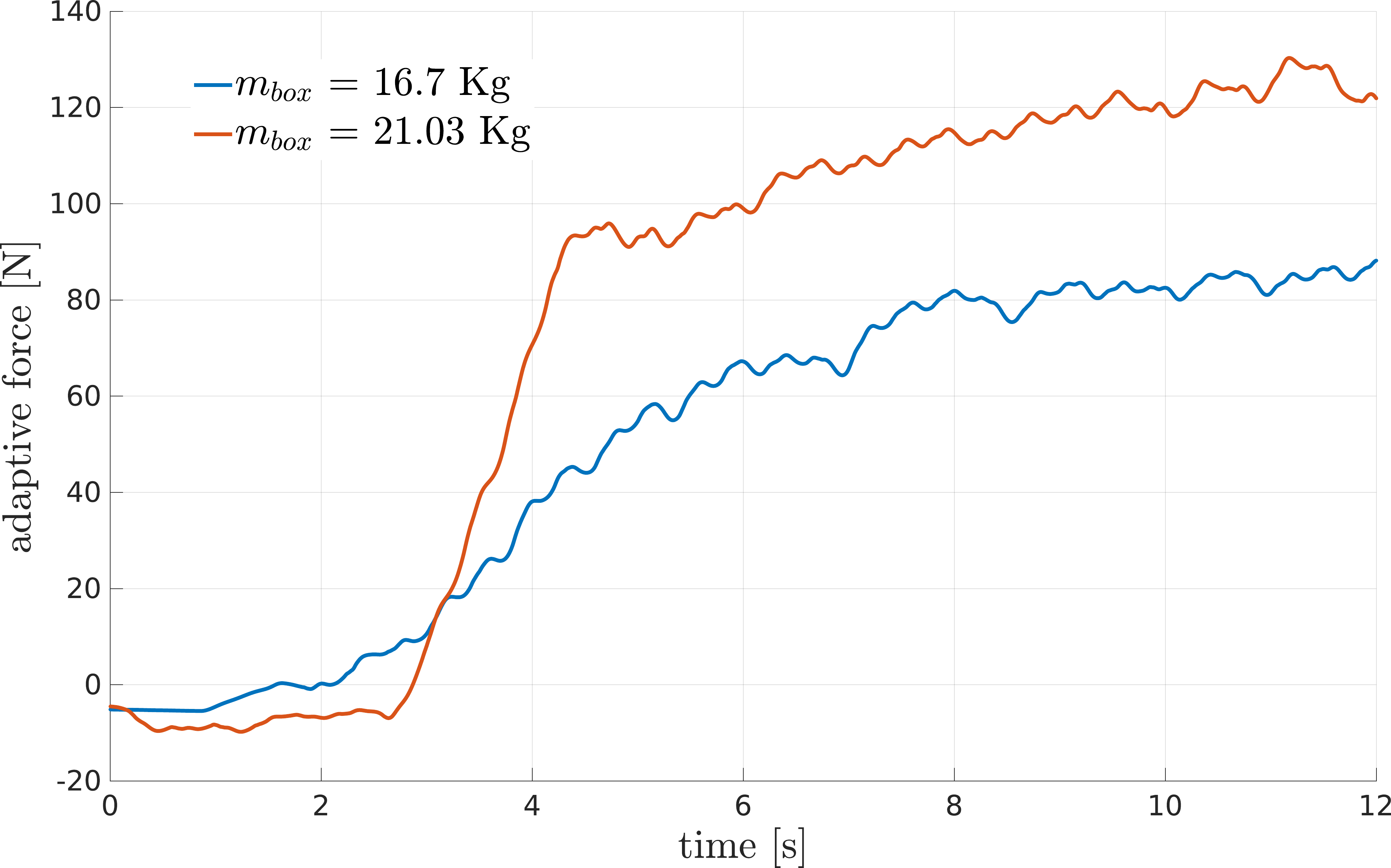}
   \caption{Adaptive force $\vY_u\hat \vpi_u$ for two experiments where the robot pulls two heavy boxes.}
\label{fig:cart_pulling_adaptive_force}    
\end{figure}

In this experiment, we are interested in making the robot pull a heavy box. We repeat the same experiment twice for two different weights of the box. In the first experiment, the box weighs 16.17 kg. In the second test, we make the task more challenging by increasing the weight up to 21.6 kg. We repeat both the experiments under the ACLF-MPC method and the nominal MPC. Base rotation and center of mass position errors are displayed in Fig. \ref{fig:cart_pulling_2_bricks_tracking_errors} and \ref{fig:cart_pulling_3_bricks_tracking_errors}, for the scenarios with the lighter box and heavier box, respectively. 
During the experiments, the desired position is computed by integrating a desired velocity of approximately 0.2 m/s, which is generated with a joystick. For safety reasons, position and orientation commands are reset whenever the errors grow larger than a threshold (for instance, for the center of mass position error along the base x-direction, the threshold is 0.1 m). As visible from Fig. \ref{fig:cart_pulling_2_bricks_tracking_errors} and \ref{fig:cart_pulling_3_bricks_tracking_errors}, the proposed formulation succeeds in improving the base tracking, and as a consequence the robot is able to pull the box forward. Also, it helps to remove oscillations that could compromise the stability of the system. This is particularly visible from the plots displaying the roll angular error and the y position error. When pulling the heavier box (Fig. \ref{fig:cart_pulling_3_bricks_tracking_errors}) such oscillations are especially strong and make the robot unstable. In Fig. \ref{fig:cart_pulling_adaptive_force}, we also show the adaptation of the wrench translational component with respect to the different friction forces that the robot experiences in the two scenarios.

\section{Conclusions and future work}
\label{sec:conclusions}
In this paper, we presented an optimal approach that unifies MPC with the online adaptation and the global stability conditions derived in adaptive control \cite{slotine1987adaptive}. We described a general formulation and an implementation for floating-base systems interacting with objects of unknown dynamic properties. We performed a number of simulations and hardware experiments on a quadrupedal robot, that demonstrated the effectiveness and the necessity of the proposed formulation.

Our implementation on a quadrupedal robot can handle external payloads and unknown constant wrenches applied on the base. In future work, we want to extend the method to adapt to extra rigid loads on the legs. In addition, a possible extension of this work would be to consider higher degree-of-freedom manipulators performing highly dynamic motions, such as catching objects, where unknown loads could be attached to any link, and the non-linear effects of more than one body would need to be adapted.

 \bibliographystyle{./bibtex/myIEEEtran} 
 \bibliography{./bibtex/IEEEabrv,bibtex/IEEEexample,bibtex/bibliography, bibtex/mybib}
\end{document}